\documentclass[10pt,conference]{IEEEtran}
\IEEEoverridecommandlockouts
\usepackage{cite}
\usepackage{amsmath,amssymb,amsfonts}
\usepackage{graphicx}
\usepackage{textcomp}
\usepackage{xcolor}
\newcommand{\textdbquote}[1]{ \textquotedblleft#1\textquotedblright~}

\usepackage{subfigure}
\usepackage{wasysym}
\usepackage[boxed,ruled,vlined,linesnumbered]{algorithm2e}
\usepackage{comment}
\usepackage{ulem}
\usepackage{makecell}
\usepackage{bm}
\usepackage{multirow}
\usepackage[colorlinks, linkcolor=red,anchorcolor=green, citecolor=blue]{hyperref}

\def\BibTeX{{\rm B\kern-.05em{\sc i\kern-.025em b}\kern-.08em
    T\kern-.1667em\lower.7ex\hbox{E}\kern-.125emX}}

\AtBeginDocument{%
  \setlength\abovedisplayskip{1ex}
  \setlength\belowdisplayskip{1ex}}
\allowdisplaybreaks[3]

\begin{document}

\title{Twin Graph-based Anomaly Detection via Attentive Multi-Modal Learning for Microservice System
\thanks{* Corresponding authors\\This work was sponsored by CCF-AFSG research fund.}
}

\author{\IEEEauthorblockN{Jun Huang$^{1}$, Yang Yang$^1$, Hang Yu$^{2*}$, Jianguo Li$^{2*}$, Xiao Zheng$^{1}$}
\IEEEauthorblockA{$^1$ \textit{School of Computer Science and Technology}, \textit{Anhui University of Technology}, Maanshan, China \\
$^2$ \textit{Ant Group}, Hangzhou, China\\
\{huangjun.cs,young978,xzheng\}@ahut.edu.cn,
\{hyu.hugo,lijg.zero\}@antgroup.com}

}

\maketitle

\begin{abstract}
Microservice architecture has sprung up over recent years for managing enterprise applications, due to its ability to independently deploy and scale services. Despite its benefits, ensuring the reliability and safety of a microservice system remains highly challenging. Existing anomaly detection algorithms based on a single data modality (i.e., metrics, logs, or traces) fail to fully account for the complex correlations and interactions between different modalities, leading to false negatives and false alarms, whereas incorporating more data modalities can offer opportunities for further performance gain. As a fresh attempt, we propose in this paper a semi-supervised graph-based anomaly detection method, MSTGAD, which seamlessly integrates all available data modalities via attentive multi-modal learning. First, we extract and normalize features from the three modalities, and further integrate them using a graph, namely MST (microservice system twin) graph, where each node represents a service instance and the edge indicates the scheduling relationship between different service instances. The MST graph provides a virtual representation of the status and scheduling relationships among service instances of a real-world microservice system. Second, we construct a transformer-based neural network with both spatial and temporal attention mechanisms to model the inter-correlations between different modalities and temporal dependencies between the data points. This enables us to detect anomalies automatically and accurately in real-time. Extensive experiments on two real-world datasets verify the effectiveness of our proposed MSTGAD method, achieving competitive performance against state-of-the-art approaches, with a 0.961 $F_1$-score and an average increase of 4.85\%.
The source code of MSTGAD is publicly available at \url{https://github.com/alipay/microservice\_system\_twin\_graph_based\_anomaly\_detection}.
\end{abstract}


\section{Introduction}
The microservice architecture has gained popularity in recent years as a method of developing applications. This methodology involves breaking down a single application into a suite of small services, each running in its own process and communicating via lightweight mechanisms~\cite{DeepTraLog:ICSE22}. One of the key advantages of this architecture is that services can be independently deployed and scaled. As a result, many industries have adopted this architecture to manage their enterprise applications. However, ensuring the reliability and safety of a Microservice system can be challenging. Traditional approaches rely on manual inspection, which is impractical for large-scale distributed systems consisting of numerous services running on different machines.

Fortunately, various types of data can be monitored in a microservice system, including service/machine metrics, logs, traces, etc. These data play a crucial role in ensuring the reliability and safety of the system. For example, metrics are real-valued time-series providing information on the status of services or machines and the associated requirements that need to be met during runtime operations.
Logs are semi-structured text messages printed by logging statements to record the system's run-time status.
Traces are hierarchical descriptions of the modules and services called upon to fulfill a user request. They are usually recorded with the service name or category and the time duration of each module. By leveraging the monitored data of a microservice system, many works 
\cite{AnomalyDetection:survey:2023, AIOps:survey:21} have successfully 
replaced manual inspection with automated anomaly detection algorithms, such as the log-based approaches \cite{DeepLog:17,LogAnomaly:IJCAI19,NeuralLog:ASE21,PLELog:ICSE2021,ICSE22:Logstudy}, metric-based approaches \cite{OCSVM:NC01,DeepSVDD:ICML18,DAGMM:ICLR18,USAD:KDD20,InterFusion:KDD21,TST:KDD21,TranAD:VLDB22}
and trace-based approaches \cite{ASE19:RepTrace,Seer:19,TraceAnomaly:ISSRE20,GMAT:ESEC/FSE20,MicroHECL:ICSE21,WWW21:MicroRank:trace,IWQOS21:TraceRCA}.

\begin{figure*}[t]
\centerline{\includegraphics[width=1\linewidth]{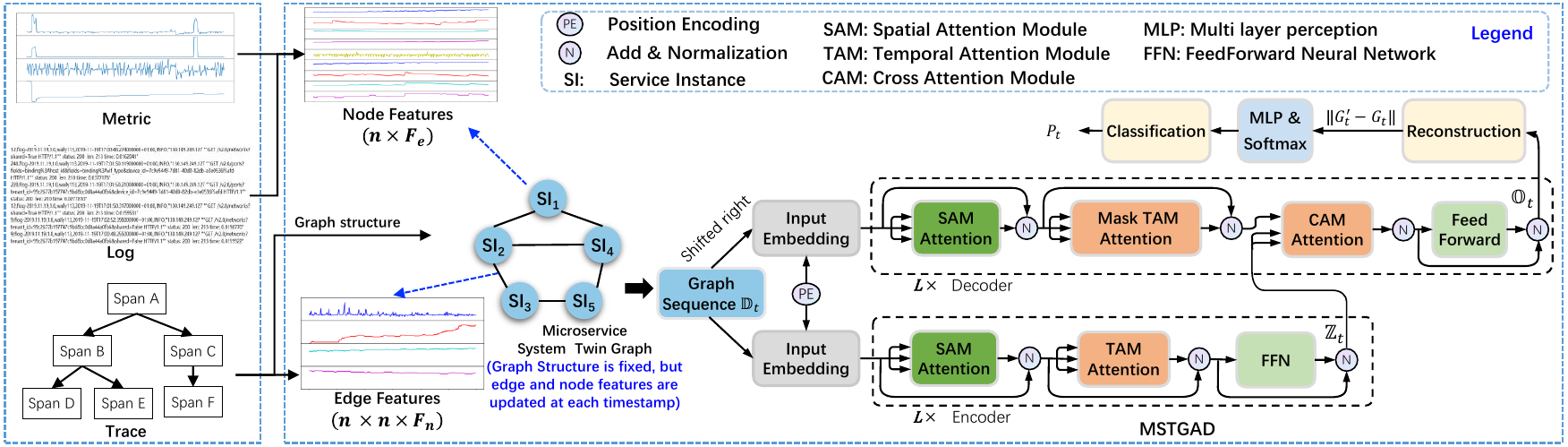}}
\caption{Framework of the proposed method MSTGAD}
\label{fig:STGAD}
\end{figure*}

Despite their effectiveness, they are often based on a single data modality, which can incur both false negatives and false alarms. Practical abnormal patterns vary across data modalities, with some only being evident in specific modalities. Failing to account for these modalities can lead to false negatives, which may further cause system outrage and substantial financial loss. 
For instance, consider an anomaly related to an API version, where a service request does not receive a timely response or is not properly parsed. Such an anomaly may not be detectable through metrics data alone, as there might not be any noticeable increase in resource consumption for the executed service container. However, it is feasible to detect this anomaly by leveraging trace data.
On the other hand, different modalities can complement each other to filter out false alarms. For example, when configuring new applications, which is a normal change, metric-based approaches may interpret increased CPU and memory usage as an anomaly, resulting in false alarms. However, by incorporating logs and traces, such false alarms can be reduced. It is essential to acknowledge that reducing the number of false alarms is of significant importance in real-world scenarios, as a high volume of false alarms can obscure true positive detections. In light of these challenges associated with single-modal anomaly detection methods, several works have proposed approaches based on multi-modal data, such as DeepTraLog~\cite{DeepTraLog:ICSE22}, SCWarn~\cite{SCwarn:ESECFSE21}, and HADES~\cite{HADES:ICSE23}. However, they use at most two modalities and do not fully consider the complex correlations between different data modalities.

Another problem with the existing works is that they are typically constructed in an unsupervised way and fail to exploit the labeling information. However, in real-world scenarios, anomaly information is usually available~\cite{Eadro:ICSE23:multimodal}, particularly when the anomaly detection algorithm fails to identify a system failure and further causes a system outage. Utilizing this labeling information can further enhance the algorithm's performance, but it cannot be incorporated into unsupervised learning methods. 
On the other hand, unlabeled data is more abundant and readily available. It is beneficial to consider a semi-supervised approach that combines both labeled and unlabeled data.

The key to remedying the above two issues lies in the extraction of useful information from heterogeneous multi-source data involved in Microservice systems and the construction of a semi-supervised model that can accurately identify anomalies. To move forward to this goal, two major challenges must be overcome.

1) \textit{How to unite and represent heterogeneous multi-source data and model the complex correlations and interactions between them?} In a Microservice system, service instances are configured in containers and called upon by user requests. Concurrently, traces are generated, and logs and metrics are monitored and recorded for these service instances and machines by timestamp. Therefore, it is natural to represent the multi-modal data using a graph to reflect the state of a Microservice system. The nodes represent service instances, and the edges correspond to their scheduling relationships. Since diverse modalities are integrated into a graph, graph machine learning techniques can be utilized to model complex correlations and interactions between them, such as the message-passing mechanism.

2) \textit{ How to construct a semi-supervised anomaly detection model?}
In the IT industry, apart from the abundant unlabeled data, some supervised information can be collected, as some anomalies can be labeled and reported by users. Therefore, the supervised information can be utilized to create an efficient semi-supervised anomaly detection model.



In this paper, we present a novel approach to address the challenges of anomaly detection in microservice systems. Our proposed method, MSTGAD, is a microservice system twin (MST) graph-based anomaly detection method that uses attentive multi-modal learning. It aims to automatically and accurately detect anomalies by leveraging metrics, logs, and traces simultaneously. The framework of MSTGAD is presented in Fig. \ref{fig:STGAD}.
We first extract and normalize the features of metrics, logs, and traces. Then, we propose to integrate the three modalities together via a graph structure, where each node represents a service instance, and the edge indicates the scheduling relationship between different service instances obtained from the traces. Since logs and metrics are recorded according to service instances, we use the features of metrics and logs to represent the node features. Traces describe the scheduling relationships between service instances within a microservice system, which are extracted and used to represent the features of edges.
Similar to a digital twin system, for each timestamp, the information about a microservice system can be reflected by the graph which can be considered a twin of the physical microservice system and virtually represent the status and scheduling relationships among service instances of a real-world microservice system.
Therefore, we name it a microservice system twin graph, i.e., MST graph. 
It can be used to design effective anomaly detection models to improve the reliability and safety of its physical counterpart.

Next, we construct a transformer-based neural network that incorporates both spatial and temporal attention mechanisms based on the MST graphs. The spatial attention module (SAM) is used to model the inter-correlation between different modalities, while the temporal attention module (TAM) is used to model the temporal dependency between data points in a sliding window. Our proposed method is evaluated on two real-world datasets, and the results demonstrate its effectiveness in anomaly detection. To sum up, our contributions include:
\begin{itemize}
    \item We perform the first empirical study of semi-supervised anomaly detection via multi-modal learning from metrics, logs, and traces simultaneously.
    
    \item We propose to integrate and represent the three types of data together via a microservice system twin (MST) graph which can be taken as a virtual representation of a real-world microservice system.
    
    \item We develop a transformer-based neural network with both spatial and temporal attention mechanisms based on the MST graphs to model the inter-correlation between different modalities and the temporal dependency among the data points (i.e., the MST graphs) of a sliding window.
    
    \item Extensive experiments on two real-world datasets show that the proposed MSTGAD model achieves competitive performance against all compared approaches, achieving 0.961 $F_1$-score with an average increase of 4.85\%. Additionally, ablation studies further validate the effectiveness of each design in our model.
\end{itemize}

\section{Related Work}
In the past decades, various techniques \cite{AnomalyDetection:survey:2023, AIOps:survey:21} have been proposed for anomaly detection based on metrics (Key Performance Indicators), logs, and traces.

Log data records the system state and significant events at various critical timestamps, which is an important and valuable resource for online monitoring, anomaly detection, and root cause analysis. There have been a lot of studies on log-based anomaly detection.
These logs-based approaches first parse the unstructured logs into structured representations through log parsing approaches including Drain \cite{Drain:ICWS17}, AEL \cite{AEL:08}, IPLoM \cite{IPLom:KDD09} and Spell \cite{Spell:ICDM16}, and SemParser \cite{SemParser:ICSE23}, then build the anomaly detection models.
Deeplog \cite{DeepLog:17} adopts LSTM to model a system log as a natural language sequence, which can automatically learn log patterns from normal execution and detect anomalies when log patterns deviate from the model trained from log data under normal execution, and also it can be incrementally updated in an online manner.
LogAnomaly \cite{LogAnomaly:IJCAI19} is an end-to-end framework also using the LSTM network to automatically detect sequential and quantitative anomalies simultaneously.
To solve log data containing previously unseen log events or log sequences, LogRobust \cite{LogRobust19} is proposed to extract semantic information of log events and represent them as semantic vectors. It then detects anomalies by utilizing an attention-based Bi-LSTM model, which has the ability to capture the contextual information in the log sequences and automatically learn the importance of different log events.
PLELog \cite{PLELog:ICSE2021} is a semi-supervised log-based anomaly detection approach via probabilistic label estimation based on an attention-based GRU neural network.
To overcome the errors caused by log parsing, NeuralLog \cite{NeuralLog:ASE21} is proposed to extract the semantic meaning of raw log messages and represent them as semantic vectors.
Existing log anomaly detection approaches treat a log as a sequence of events and cannot handle microservice logs that are distributed in a large number of services with complex interactions.

Inspired by kernel-based one-class classification \cite{OCSVM:NC01}, Deep SVDD \cite{DeepSVDD:ICML18} is proposed to train a neural network by minimizing the volume of a hypersphere that encloses the network representations of the data.
Tuli \textit{et. al.} propose a transformer-based anomaly detection model TranAD \cite{TranAD:VLDB22} which uses self-conditioning and an adversarial training process.
The DAGMM \cite{DAGMM:ICLR18} method uses a deep autoencoding Gaussian mixture model for dimension reduction in the feature space and recurrent networks for temporal modeling.
The USAD \cite{USAD:KDD20} method uses an autoencoder with two decoders with an adversarial game-style training framework. This is one of the first works that focus on low overheads by using a simple autoencoder and can achieve a several-fold reduction in training times compared to the prior art. The GDN \cite{GDN:AAAI21} approach learns a graph of relationships between data modes and uses attention-based forecasting and deviation scoring to output anomaly scores. 
In microservice systems, traces are widely used in anomaly detection and root cause analysis.
Seer \cite{Seer:19} is an online cloud performance debugging system that leverages deep learning and a massive amount of tracing data to learn spatial and temporal patterns to detect Qos violations.
TraceAnomaly \cite{TraceAnomaly:ISSRE20} is an
unsupervised anomaly detection approach of microservice through Service-Level Deep Bayesian Networks.
GMTA \cite{GMAT:ESEC/FSE20} is a graph-based approach  for architecture understanding and problem diagnosis of microservice trace analysis.
Liu \textit{et al.} propose a high-efficient root cause localization approach MicroHECL \cite{MicroHECL:ICSE21} which dynamically constructs a service call graph and analyzes possible anomaly propagation chains by traversing the graph along anomalous service calls.

Although the above approaches have achieved great performance in anomaly detection. While they are mainly constructed based on single-modal data, such as logs, metrics, or traces respectively.
However, as a single source of information is often insufficient to depict the status of a microservice system precisely, existing methods could produce many false alarms and may omit some true positives \cite{HADES:ICSE23}.
Recently, several studies have been done based on multi-modal data to further improve the performance of anomaly detection.

DeepTraLog \cite{DeepTraLog:ICSE22} utilizes a unified graph representation to describe the complex structure of a trace together with log events embedded in the structure, and trains a GGNNs-based deep SVDD \cite{DeepSVDD:ICML18} model and detects anomalies in new traces and the corresponding logs.
SCWarn \cite{SCwarn:ESECFSE21} is proposed for online service systems to identify the bad software changes via multimodal learning from heterogeneous multi-source data. 
In SCWarn, the temporal dependency in each time series is captured by the LSTM model, and the inter-correlations among multi-source data are encoded via multimodal fusion.
Similar to SCWarn, Liu \textit{et al.} propose HADES \cite{HADES:ICSE23} 
which employs a hierarchical architecture to learn a global representation of the system status by fusing log semantics and metric patterns, and a novel cross-modal attention module to capture discriminative features and meaningful interactions from multi-modal data.

In parallel to our research, several other endeavors have emerged to explore the combined utilization of logs, metrics, and traces as modalities, such as \cite{Eadro:ICSE23:multimodal,AnoFusion,DiagFusion}
Specifically, Eadro~\cite{Eadro:ICSE23:multimodal} presents an end-to-end framework that seamlessly integrates anomaly detection and root cause localization based on multi-source data, specifically designed for troubleshooting purposes. This framework adeptly models both intra-service behaviors and inter-service interactions, exploiting shared knowledge through multi-task learning. AnoFusion~\cite{AnoFusion} offers an unsupervised failure detection approach that efficiently identifies instance failures through multimodal data. It employs a Graph Transformer Network (GTN) to effectively capture correlations within the heterogeneous multimodal data. Additionally, it seamlessly integrates a Graph Attention Network (GAT) with a Gated Recurrent Unit (GRU) to address the challenges introduced by dynamically changing multimodal data. DiagFusion~\cite{DiagFusion}, on the other hand, employs embedding techniques and data augmentation to accurately represent multimodal data of service instances. By combining deployment data and traces, it constructs a dependency graph and subsequently employs a graph neural network to localize the root cause instance and determine the type of failure. It is important to note that all the aforementioned approaches are either fully supervised or unsupervised. In contrast, our proposed method is semi-supervised, which effectively leverages both labeled and unlabeled data to enhance model performance. Furthermore, the existing works consider all three modalities as node features in their graph neural networks for capturing spatial dependence, but our approach treats traces as edge features. Additionally, Eadro~\cite{Eadro:ICSE23:multimodal} and AnoFusion~\cite{AnoFusion} employs CNN and GRU to capture the temporal dependence, whereas our proposed method harnesses the power of Transformers. Transformers have been proven to outperform CNNs and GRUs for time series modeling~\cite{li2019enhancing,liu2021pyraformer}. 
Lastly, it is worth mentioning that AnoFusion primarily focuses on anomaly detection, DiagFusion on root cause analysis, while Eadro tackles both anomaly detection and root cause localization. In line with our work, our objective here is anomaly detection, akin to AnoFusion, but distinct from DiagFusion and Eadro.
\section{Preliminaries}
In our proposed model, MSTGAD, we use various attention mechanisms to capture the dependencies between different parts of the input data. Before delving into the details of our model, we first introduce different attention mechanisms used in the following sections.


\subsection{Scaled Dot-Product Attention}
The fundamental concept of attention is to determine a series of weights that express the significance or pertinence of various elements within the input information, depending on a particular inquiry. These weights can be utilized to calculate a weighted sum of the input data, which may then be utilized as an input for the subsequent layer of the model. Specifically, given the query $\mathbf{Q} \in \mathbb{R}^{L_Q \times d_k}$, key $\mathbf{K}\in \mathbb{R}^{L_K \times d_k}$, and value $\mathbf{V}\in \mathbb{R}^{L_V \times d_v}$,
a scaled-dot product attention \cite{vaswani_attention_2017} of the three matrices can be defined as
\begin{equation}\label{eq:attention}
    \text{Attention}(\mathbf{Q},\mathbf{K},\mathbf{V})=\text{Softmax}(\frac{\mathbf{QK}^\top}{\sqrt{d_k}})\mathbf{V},
\end{equation}
where $\text{Softmax}(\mathbf{Q} \mathbf{K}^\top / \sqrt{d_k}) \in \mathbb{R}^{L_Q \times L_K}$ can be considered as a similarity or attention score matrix, and $d_k$ is the number of features. By selectively focusing on different parts of the input data, attention can help the model to better capture long-term dependencies and improve its overall performance.

\subsection{Multi-head Attention}
Multihead attention is an extension of the above standard attention mechanism that allows the model to jointly attend to information from different representation subspaces. The basic idea behind multi-head attention is to split the input queries, keys, and values into multiple heads, and to compute separate attention scores for each head. The outputs from each head are then concatenated and passed through a linear projection to generate the final output. By using multiple heads, the model can learn to attend to different aspects of the input data, and the final output is a combination of the outputs from multiple heads. Mathematically, the multi-head attention function can be expressed as follows:
\begin{align}\label{eq:multihead}
    \text{MultiHead}(\mathbf{Q},\mathbf{K},\mathbf{V})&= \text{Concat}(H_1, \dots, H_h)\,\mathbf{W}^O, 
\end{align}
where each $H_i=\text{Attention}(\mathbf{Q}\mathbf{W}_i^Q,\mathbf{K}\mathbf{W}_i^K,\mathbf{V}\mathbf{W}_i^V)$ is single-head attention, and $h$ is the number heads.
The projections $\mathbf{W}_i^Q \in \mathbb{R}^{d_{model} \times d_k} $, $\mathbf{W}_i^K \in \mathbb{R}^{d_{model} \times d_k} $ , $\mathbf{W}_i^V \in \mathbb{R}^{d_{model} \times d_v} $ and $\mathbf{W}^O \in \mathbb{R}^{hd_v \times d_{model}}$ are learnable parameters.

\subsection{Graph Attention}
Graph attention is also an extension of the scaled dot-product attention mechanism that can be used to process graph-structured data. Unlike traditional attention mechanisms where the input is a sequence of vectors, in graph attention, the input is a graph where each node represents a vector. Each node is associated with \textdbquote{query},  \textdbquote{key}, and  \textdbquote{value} vectors, which are used to compute attention scores between pairs of nodes. Specifically, For a graph $G = (V, A, E)$, $V$ is the set of nodes, $E$ is the set of edges and $A$ is the adjacency matrix. A Graph Attention Network (GAT)~\cite{gat,GATv2,egat} layer updates each node representation by aggregating the representations of its neighboring nodes. Each node $v_i$ can be updated by
\begin{equation}
    v_i^{'} = \sum_{u \in \mathbb{N}(i)}\alpha_{i, u}  \mathbf{W}v_u,
\end{equation}
where $\mathbf{W}\in  \mathbb{R}^{ d_{model}\times F_v}$ is a learnable parameter, and the graph attention weight $\alpha_{ij}$ is defined as
\begin{align}
    \alpha_{i, j} &= \frac{e(v_i, v_j, e_{ij})}{\sum_{u \in \mathbb{N}(i)}e(v_i, v_u,e_{iu})},
\end{align}
where $v_i\in \mathbb{R}^{F_v}$ and $v_j\in \mathbb{R}^{F_v}$ are the $i$-th and $j$-th nodes' feature representation, and $e_{ij}$ is the edge representation between the two nodes.
Different from standard GAT, the edges are also used to calculate the attention weight.
$\mathbb{N}(i)$ includes the $i$-th node and its directed connected neighbors.
$e(v_i, v_j, e_{ij})$ indicates the importance of the features of node $j$ to node $i$, which is defined as
\begin{align}
    e(v_i, v_j, e_{ij}) &= \beta^\top \text{LeakyReLU}(\mathbf{W} [v_i \|  v_j \|  e_{ij}]),
\end{align}
where ${\beta}\in \mathbb{R}^{3d_{model}}$ and $ \mathbf{W} \in \mathbb{R}^{3d_{model}\times (2F_v+F_e)}$ are learnable parameters, and $\|$ denotes vector concatenation.
Equivalently, the above formulations can be written in the matrix form as
\begin{align}
    \label{eq:spatialAtt}
\text{SpatialAtt}(\mathbf{V},\mathbf{V}, \mathbf{E}) &= \text{Softmax}(e(\mathbf{V}, \mathbf{V}, \mathbf{E})) \mathbf{V}\mathbf{W}^\top, \\
    \text{where}~ e(\mathbf{V},\mathbf{V}, \mathbf{E}) &= {\bm{\beta}}^\top \text{LeakyReLU}( \mathbf{W}[\mathbf{V} \|  \mathbf{V} \|  \mathbf{E}]).\nonumber
\end{align}

\section{Approach}
The proposed method, MSTGAD, aims to automatically and accurately detect anomalies by leveraging multi-modal learning from metrics, logs, and traces simultaneously. The approach takes all three modalities as input and trains a graph-based deep learning model based on an enhanced transformer structure.
The framework of MSTGAD is presented in Fig. \ref{fig:STGAD}, which is mainly composed of two stages, i.e., multi-modal data fusion and representation, and anomaly detection.
In the first stage, the features of metrics, logs, and traces are extracted and integrated using an MST graph. The graph represents service instances as nodes and scheduling relationships between different instances as edges. In the second stage, a transformer-based neural network is constructed with spatial and temporal attention based on the MST graphs, for the sake of anomaly detection. The network can model the inter-correlation between different modalities and the temporal dependency of data points in a sliding window. Next, we elaborate on these two stages.

\subsection{Pre-Processing and MST graph construction}
\subsubsection{Metric Pre-processing}
Metrics provide valuable information regarding the status of services and machines, such as response time, the number of threads, and CPU and memory usage. In this paper, we focus on multivariate time-series data, denoted as $\mathbb{T}_{Metric}=\{\mathbf{M}_1,\mathbf{M}_2,...,\mathbf{M}_T\}$, which represents a sequence of timestamped observations of size $T$. Each data point $\mathbf{M}_t\in\mathbb{R}^{N \times F_m}$ is collected at a specific timestamp $t$ for $N$ service instances and machines, where $F_m$ is the number of metrics. We further normalize the data and convert it to time-series windows for both training and testing purposes. Specifically, each data point $\mathbf{M}_t$ is normalized using min-max normalization, as follow:
\begin{equation}
    \mathbf{M}_t = \frac{\mathbf{M}_t-\min(\mathbb{T}_{Metric})}{\max(\mathbb{T}_{Metric})-\min(\mathbb{T}_{Metric})+\epsilon},
\end{equation}
where $\max(\mathbb{T}_{Metric})\in\mathbb{R}^{ F_m}$ and $\min(\mathbb{T}_{Metric})\in\mathbb{R}^{ F_m}$ are the maximum and minimum vectors in the training time-series, and $\epsilon$ is a small constant vector used to avoid zero-division.

In practice, the number of metrics might be huge. To cope with the challenge of real-time data collection, we can remove those metrics whose variances are very low, since anomalies are reflected by metrics with fluctuation.

\subsubsection{Log Parsing}
Logs are text messages that are semi-structured and record system states and significant events at critical timestamps. In this step, we aim to convert unstructured log messages into structured log events. To achieve this, we use the widely adopted parser, Drain3~\cite{Drain:ICWS17}, similar to other studies on anomaly detection~\cite{DeepTraLog:ICSE22,HADES:ICSE23,SCwarn:ESECFSE21}.
Drain3 can parse logs in a streaming and timely manner with high parsing accuracy and efficiency. After parsing, timestamp, service instance ID, and log template index are attached to each log event for further analysis. To enable combined analysis with metrics, we count the occurrences of each template in each timestamp, creating log representations consistent with the formulation of metrics. This yields a timestamped log sequence, denoted as $\mathbb{T}_{Log}=\{\mathbf{L}_1,\mathbf{L}_2,...,\mathbf{L}_T\}$ of observation of size $T$, where each data point $\mathbf{L}_t\in \mathbb{R}^{N\times F_l}$, and $F_l$ is the number log templates.

Similar to the normalization of metrics, each data point $\mathbf{L}_t$ is also normalized via the min-max normalization as
\begin{equation}
    \mathbf{L}_t = \frac{\mathbf{L}_t-\min(\mathbb{T}_{Log})}{\max(\mathbb{T}_{Log})-\min(\mathbb{T}_{Log})+\epsilon},
\end{equation}
where $\max(\mathbb{T}_{Log})\in\mathbb{R}^{ F_l}$ and $\min(\mathbb{T}_{Log})\in\mathbb{R}^{ F_l}$ are the maximum and minimum vectors in the training time-series. 

Serializing logs can potentially introduce challenges associated with long-tail distributions. However, it is possible to mitigate this issue to some extent by leveraging the complementary information from other modalities. For instance, although the occurrence of out-of-memory anomalies is rare and may be affected by the long-tail problem, such anomalies can also be detected through the analysis of metric data. Furthermore, the performance of our method can be further enhanced by incorporating more advanced log parsing techniques~\cite{TKDE23:logparingSurvey}, including semantic-based approaches~\cite{NeuralLog:ASE21,SemParser:ICSE23}.

\begin{figure}[t]    \centerline{\includegraphics[width=0.9\linewidth]{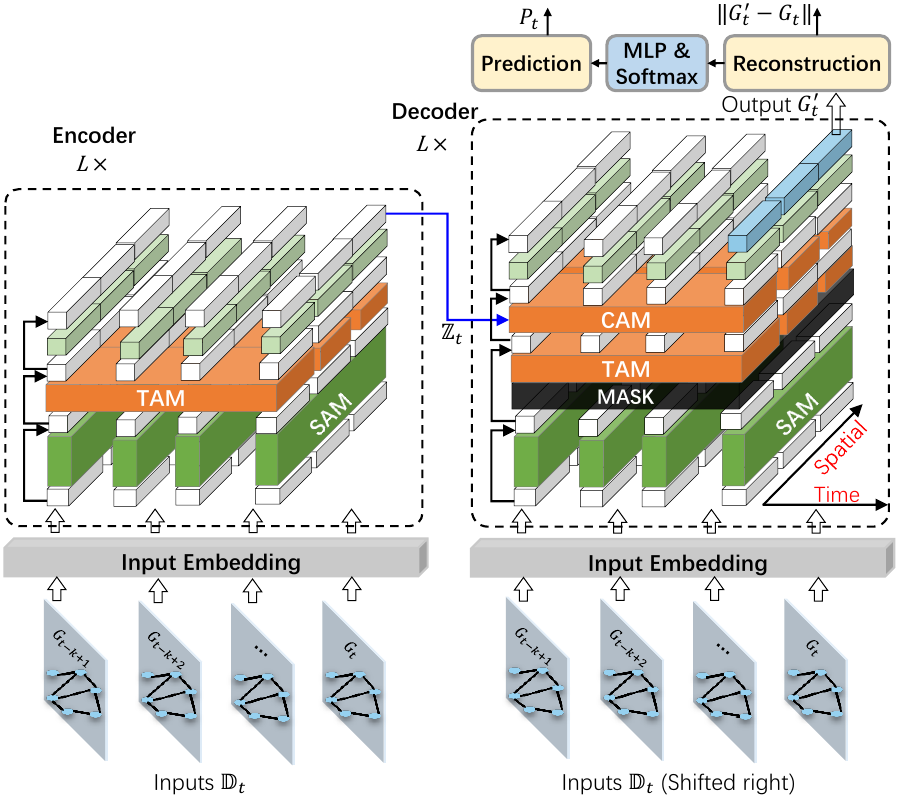}}
    \caption{Framework of MSTGAD}
    \label{fig:MSTGAD}
\end{figure}
\subsubsection{Trace Parsing}
Each trace in a microservice system depicts the execution processes of user requests. These processes are known as spans, and they possess valuable attributes such as trace ID, request type, span ID, father span ID, service instance ID, start time, and duration time. In the spirit of TraceAnomaly~\cite{TraceAnomaly:ISSRE20}, we represent spans as time series. 
Within a sliding window, we aggregate the total duration time of spans sharing the same request type, service instance launcher, and service instance receiver for each timestamp. It is important to note that unfinished spans are omitted from the window.
Consequently, we obtain a timestamped trace sequence $\mathbb{T}_{Trace}=\{\mathbf{S}_1,\mathbf{S}_2,...,\mathbf{S}_T\}$ of size $T$, and each data point $\mathbf{S}_t\in\mathbb{R}^{N\times N\times F_s}$, where $N$ indicates the number of service instances of a microservice system, and $F_s$ indicates the dimension of the extracted features of traces.
Since not all service instances are interconnected, we use an approximate normalization technique to differentiate between small spans and no spans. Specifically, we define it as:
\begin{equation}
    \mathbf{S}_t = \frac{\mathbf{S}_t}{\text{mean}(\mathbb{T}_{Trace})},
\end{equation}
where $\operatorname{mean}(\mathbb{T}_{Trace})\in\mathbb{R}^{F_s}$ is the average of $\mathbb{T}_{Trace}$

In reality, if tremendous traces are generated, state-of-the-art head-based and tail-based sampling techniques can be applied to reduce the processing and storage costs.


\subsubsection{Graph Construction}
\label{section:graphconstruction}
After obtaining $\mathbb{T}_{Metrics}$, $\mathbb{T}_{Log}$, 
and $\mathbb{T}_{Trace}$, we can integrate these three modalities together by constructing an MST graph to virtually represent the status and scheduling relationships among service instances of a real-world microservice system.
In an MST graph, nodes represent service instances, and edges represent scheduling relationships between different instances.

Formally, for each timestamp $t$, the MST graph is defined as $G_t=<V_t, A_t, E_t>$, where $V_t$ is the set of nodes and the corresponding node features are defined as $\mathbf{V}_t=\mathbf{M}_t\|\mathbf{L}_t$, i.e., the concatenation of metric and log features, and $\|$ indicates the concatenation operator.
$E_t$ is the set of edges, and $\mathbf{E}_t=\mathbf{S}_t$ is used to represent the features of edges, and $A_t$ is the adjacent matrix.
Each element $a_{ij}$ of $A_t$ is defined as
\begin{equation}\nonumber
    \begin{array}{l}
    a_{ij}
    =\left\{ \begin{array}{l}
    1, ~~~\text{if their is scheduling between the $i$-th}\\
    \quad\quad\text{and $j$-th service instances}\\
    0,~~~\text{otherwise}
    \end{array} \right.
    \end{array}.
\end{equation}

In reality, microservice systems undergo constant changes, such as services scaling up or down. These changes can be accommodated by updating the adjacency matrix when services are added or removed. However, it is important to note that the parameters of the Graph ATtention network (GAT) used in the proposed model, which will be introduced in the following sections, can remain unchanged. This is because the GAT operates on the graph structure itself, rather than being dependent on the specific services present in the system. As a result, the model can effectively adapt to changes in the microservice system without the need for retraining or modifying the GAT parameters.
Furthermore, by considering the preprocessed metrics and logs as node features, and the traces as edge features, we can characterize the dynamic behavior of traces by adjusting the edge features accordingly, without modifying the underlying graph structure. This property allows for efficient representation of changes in the scheduling relationships among service instances. Consequently, when a service is removed from the system, we can alternatively set the corresponding node and edge features to zero without removing the service from the graph.

In summary, the real-time status and scheduling relationships among service instances of a real-world microservice system can be virtually represented by the MST graph, which can be used to design an effective anomaly detection model to improve the reliability and safety of its physical counterpart.

To model the temporal dependence of a data point at timestamp $t$, we consider a sliding window of length $k$ as
\begin{equation}
    \mathbb{D}_t = \{G_{t-k+1},\cdots,G_{t}\}. \label{eq:D_t}
\end{equation}
For each sliding window $\mathbb{D}_t$, $y_t\in\mathbb{R}^{1\times N}$ is the ground truth label vector, and $y_{ti}\in\{-1,0,1\}$ is the ground truth label for the $i$-th service instance in $G_{t}$.
The numbers $1$ and $0$ indicate abnormal and normal data respectively, and $-1$ indicates the corresponding label is unknown for the unlabelled data.

\subsection{Transformer with Spatial and Temporal Attention}
\label{sec:MSTGAD}
MSTGAD is constructed based on the transformer \cite{vaswani_attention_2017} architecture for anomaly detection via attentive multi-modal learning.
Fig. \ref{fig:MSTGAD} shows the architecture of the neural network used in MSTGAD, which consists of encoding (left side) and decoding (right side) steps with several attention modules.

In the encoding step, the encoder takes the input sequence $\mathbb{D}_t$ with $k$ MST graphs (cf.~Eq~\eqref{eq:D_t}) and converts it to a hidden state $\mathbb{Z}_t$. To accomplish this, $\mathbb{D}_t$ is first subjected to the Input Embedding operation, a linear layer that maps multi-source features to appropriate dimensions for multi-head attention. We also add position encoding~\cite{vaswani_attention_2017} of each data point on the time dimension to provide MSTGAD with local position information.
In order to model the inter-correlation between different modalities, we propose the spatial attention module (SAM), which is based on GAT~\cite{GATv2}.
Moreover, we propose the temporal attention module (TAM) to model the temporal dependency between different data points in the sliding window $\mathbb{D}_t$. Specifically, the workflow of each layer of the encoder can be defined by a series of operations:
\begin{align}
    EI_{1} &= \text{Norm}(EI_{0} + \text{SAM}(EI_{0})), \nonumber \\
    EI_{2} &= \text{Norm}(EI_{1} + \text{TAM}(EI_{1})), \nonumber\\
    \mathbb{Z}_{t} &= \text{Norm}(EI_{2} + \text{FFN}(EI_{2})),\nonumber
\end{align}
where $\text{Norm}(\cdot)$ is the normalization operation, and FFN is the Feed-forward neural network.

In the decoding step, $\mathbb{D}_{t-1}$ will be used to predict the representation of $\mathbb{D}_t$.
To achieve this goal, $\mathbb{D}_t$ is shifted one timestamp to the right and padded with zero at the first timestamp as the input of decoding. After that, the input sequence $\mathbb{D}_{t-1}$ is also converted to $DI_0$ via the Input Embedding operation.
The spatial attention module (SAM) is then applied to model the inter-correlation between different modalities. Unlike the encoding step, TAM is run with a causal mask to ensure that each $G_t$ is updated or predicted only based on its former data points. In addition, we propose a cross-attention module (CAM) to update $\mathbb{D}_t$ with the guidance of $\mathbb{Z}_t$. Correspondingly, the workflow of each layer of the decoder can be defined as the following operations:
\begin{align}
    DI_{1} &= \text{Norm}(DI_{0} + \text{SAM}(DI_{0})), \nonumber \\
    DI_{2} &= \text{Norm}(DI_{1} + \text{Mask}(\text{TAM}(DI_{1}))), \nonumber \\
    DI_{3} &= \text{Norm}(DI_{2} + \text{CAM}(DI_{2}, \mathbb{Z}_t)), \nonumber \\
    \mathbb{O}_t &= \text{Norm}(DI_{3} + \text{FFN}(DI_{3})). \nonumber
\end{align}

Finally, we detect anomalies based on the reconstruction errors between $\mathbb{D}_t$ and $\mathbb{O}_t$. 
The following subsections provide further details about the SAM, TAM, and CAM attention modules.
$\mathbb{D}_t$ is the input of MSTGAD, 
and $\mathbf{M}_{t-k+1:t}$, $\mathbf{L}_{t-k+1:t}$, and $\mathbf{S}_{t-k+1:t}$ are used to represent the features of metrics, logs, and traces of $\mathbb{D}_t$ respectively for convenience.
\begin{figure}[t]
        \subfigure[Update the feature of nodes]{
        \includegraphics[width=0.2\textwidth,height=0.9in]{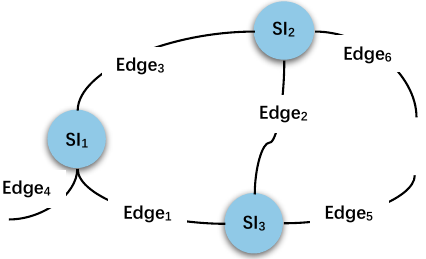}}\hspace{.2in}
        \subfigure[Update the feature of edges]{
        \includegraphics[width=0.2\textwidth,height=0.8in]{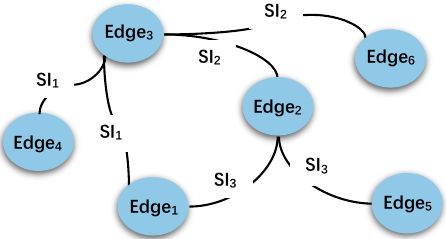}} 
    \caption{Spatial Attention}
    \label{fig:SpatialAtt}
\end{figure}

\subsubsection{Spatial Attention Module}
As mentioned in the introduction, metrics, logs, and traces are different ways of viewing the status of a microservice system. By exploiting the complementary information from these different modalities, we can boost the generalization ability of anomaly detection algorithms. To achieve this, we propose a spatial attention module (SAM) that models the inter-correlation between modalities using two layers of Graph Attention Network (GAT)~\cite{GATv2}.

SAM allows for the updating of node and edge features by incorporating information from their neighbors through message passing. Fig. \ref{fig:SpatialAtt} provides a toy example, where SI stands for service instance.
In Fig. \ref{fig:SpatialAtt}(a), when updating the representation of a node, the information from its directed neighbors will be passed to it with an attention weight along the edges.
Similarly, the feature of edges can be updated by exchanging the roles of nodes and edges in an MST graph (see Fig. \ref{fig:SpatialAtt}(b)).
As a result, the representation of each modal can be updated and improved by absorbing the complementary information from other modalities.

In this paper, for an input MST graph $G_t=(V_t,A_t, E_t)$, we apply multi-head attention (see Eq.(\ref{eq:multihead})), and then the features of nodes of service instances of $G_t$ can be updated by
\begin{align}
    \mathbf{V}_t^{'} = \text{MultiHead}(\mathbf{V}_t,\mathbf{V}_t, \mathbf{E}_t),
\end{align}
where each head $H_i=\text{SpatialAtt}(\mathbf{V}_t,\mathbf{V}_t, \mathbf{E}_t)$ (see Eq.(\ref{eq:spatialAtt})).
Unlike standard GAT~\cite{GATv2}, our approach incorporates edges in the calculation of the attention weight. The reasoning behind this is that edge features represent the features of traces, and excluding them would result in the loss of valuable trace information.
By exchanging the roles of nodes and edges in an MST graph, the features $\mathbf{E}_t$ of edges of service scheduling in each $G_t$ can be updated in the same manner as
\begin{align}
    \mathbf{E}_t^{'} = \text{MultiHead}(\mathbf{E}_t,\mathbf{E}_t, \mathbf{V}_t^{'}),
\end{align}
where each head $H_i=\text{SpatialAtt}(\mathbf{E}_t,\mathbf{E}_t, \mathbf{V}_t^{'})$.

The SAM module consists of two layers of GAT which alternate in updating the features of nodes and edges. This enables the module to efficiently model the complex inter-correlation among metrics, logs, and traces by learning the representation of each MST graph $G_t$ in a sliding window $\mathbb{D}_t$. 
Through the thorough integration and fusion of complementary information among the three modalities, the SAM module produces an enriched representation $G_t^{'}$. 
Thus, for a sliding window, we will update all the MST graphs to get a new representation of $\mathbb{D}_t$ as
\begin{align}
    \mathbb{D}_t^{\text{SAM}}=(G_{t-k+1}^{'},\dots,G_{t}^{'}) = \text{SAM}(\mathbb{D}_t). \nonumber
\end{align}
\vspace{-10pt}
\subsubsection{Temporal Attention Module}
\begin{figure}[t]
    \centerline{\includegraphics[width=0.8\linewidth]{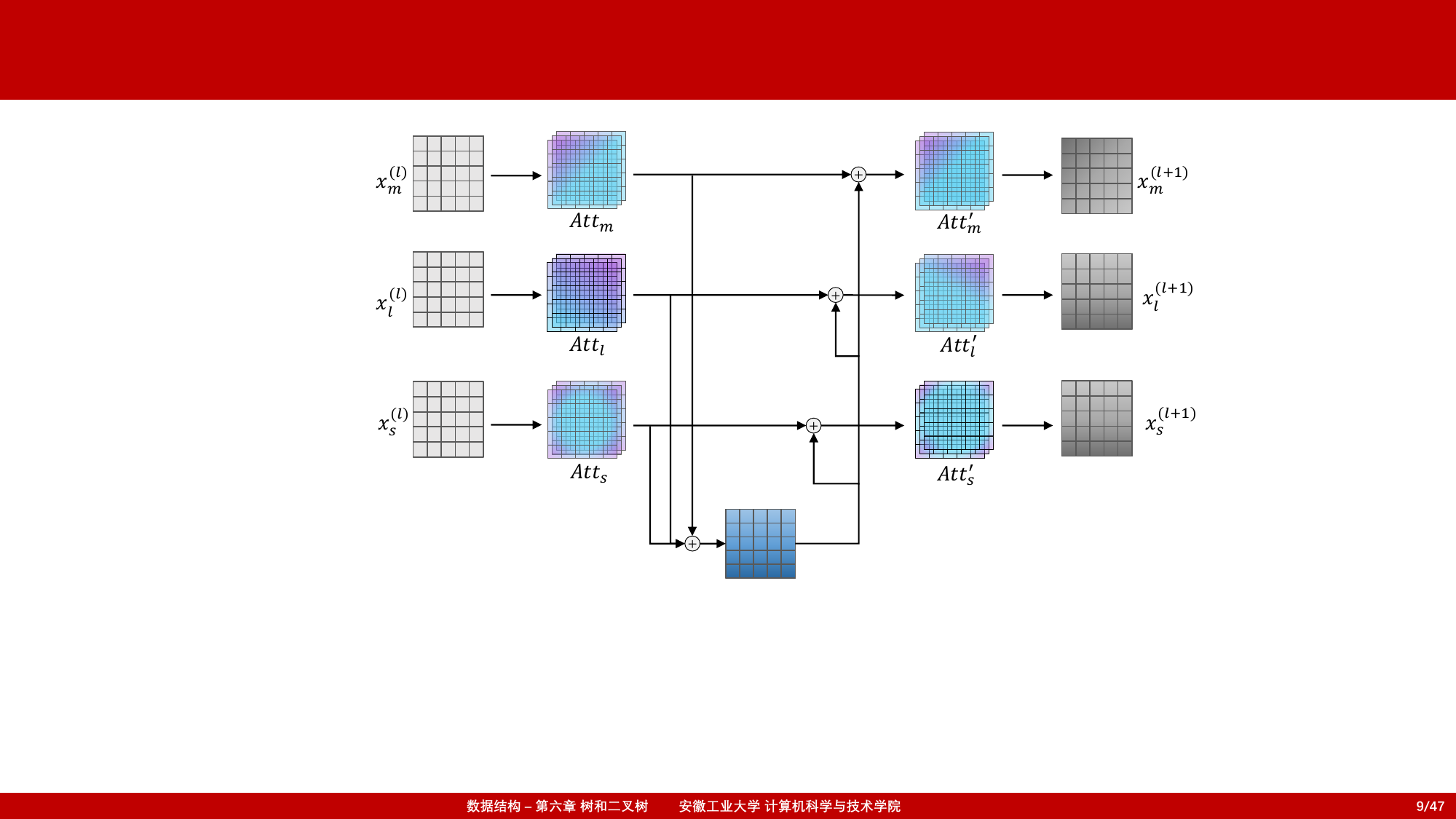}}
    \caption{Framework of the Temporal Attention}
    \label{fig:TemporalAtt}
\end{figure}

The above SAM module is designed to model the complex inter-correlation among different modalities at the local graph level across space. In this section, we further propose the temporal attention module (TAM) to capture the complex inter-correlation among different modalities at the global sequence level across time. Notice that in a microservice system, multi-source monitored data are generated sequentially and exhibit temporal trends. To improve anomaly detection performance, it is crucial to model temporal dependencies within these sequences. Therefore, we adopt the TAM module to capture time dependencies for data points in a sliding window.
As a first step, we extend the attention mechanism in Eq.~(\ref{eq:attention}) as
\begin{equation}\label{eq:temporalAtt}
    \text{TemporalAtt}(\mathbf{Q},\mathbf{K},\mathbf{V},\mathbf{C})=\text{Softmax}(\frac{\mathbf{Q}\mathbf{K}^\top}{\sqrt{d_k}}+\mathbf{C})\mathbf{V},
\end{equation}
where $\mathbf{C}$ is the average of attention scores of multi-source data. This is used to capture the common temporal dependencies shared by the multi-source data and fuse the complementary information among them.

Next, we apply the multi-head self-attention across time to a sequence of the spatial representation of the MST graph given by the SAM. The structure of TAM is shown in Fig. \ref{fig:TemporalAtt}. 
By capturing the temporal dependency, the feature of metrics $\mathbf{M}_{t-k+1:t}$ can be updated as
\begin{align}\nonumber
    \mathbf{M}_{t-k+1:t}^{'} =\text{Multihead}(&\mathbf{M}_{t-k+1:t},\mathbf{M}_{t-k+1:t},\mathbf{M}_{t-k+1:t}
    ,\mathbf{C}),\nonumber
\end{align}
where each $H_i=\text{TemporalAtt}(\mathbf{M}_{t-d+1:t}\mathbf{W}_i^Q, \mathbf{M}_{t-d+1:t}\mathbf{W}_i^K,$ $\mathbf{M}_{t-d+1:t}\mathbf{W}_i^V,\mathbf{C})$, and the average of attention score $\mathbf{C}$ is defined as
\begin{scriptsize}
\begin{equation}
    (\frac{{\mathbf{M}_{t-k+1:t}}{\mathbf{M}_{t-k+1:t}^{\top}}}{\sqrt{F_m}} +\frac{{\mathbf{L}_{t-k+1:t}}{\mathbf{L}_{t-k+1:t}^{\top}}}{\sqrt{F_l}} +\frac{{\mathbf{S}_{t-k+1:t}}{\mathbf{S}_{t-k+1:t}^{\top}}}{\sqrt{F_s}})/3. \nonumber
\end{equation}
\end{scriptsize}
Similarly, the feature of logs and traces can be updated as
\begin{align}\nonumber
    \mathbf{L}_{t-k+1:t}^{'} =\text{Multihead}(&\mathbf{L}_{t-k+1:t},\mathbf{L}_{t-k+1:t},\mathbf{L}_{t-k+1:t}
    ,\mathbf{C}),\nonumber\\
    \mathbf{S}_{t-k+1:t}^{'} =\text{Multihead}(&\mathbf{S}_{t-k+1:t},\mathbf{S}_{t-k+1:t},\mathbf{S}_{t-k+1:t}
    ,\mathbf{C}).\nonumber
\end{align}
In summary, we will update all modalities within $\mathbb{D}_t$ as
\begin{align}\nonumber
    \mathbb{D}_t^{\text{TAM}} &= \text{TAM}(\text{Norm}(\mathbb{D}_t+\mathbb{D}_t^{\text{SAM}}))\\ \nonumber
    &=(\mathbf{M}_{t-k+1:t}^{'},\mathbf{L}_{t-k+1:t}^{'},\mathbf{S}_{t-k+1:t}^{'}).
\end{align}
\vspace{-10pt}
\subsubsection{Cross Attention Module}
The cross-attention module CAM is proposed to perform decoding using the hidden state $\mathbb{Z}_t$ given by the encoder. This module assists the decoder in obtaining guidance information from $\mathbb{Z}_t$.
To achieve this goal, we adopt the multi-head attention (see Eq.(\ref{eq:multihead})).
To distinguish features between $\mathbb{D}_t$ and $\mathbb{Z}_t$, we add a superscript, e.g. $\mathbf{M}^{\mathbb{Z}}_{t-k+1:t}$ and $\mathbf{M}^{\mathbb{D}}_{t-k+1:t}$, to indicate the feature of metrics in the encoding and decoding stages respectively.
In CAM, features of the three modalities are updated by 
\begin{align}
    \mathbf{M}^{'\mathbb{D}}_{t-k+1:t} &= \text{MultiHead}(\mathbf{M}^{\mathbb{D}}_{t-k+1:t},\mathbf{M}^{\mathbb{Z}}_{t-k+1:t},\mathbf{M}^{\mathbb{Z}}_{t-k+1:t}), \nonumber\\
    \mathbf{L}^{'\mathbb{D}}_{t-k+1:t} &= \text{MultiHead}(\mathbf{L}^{\mathbb{D}}_{t-k+1:t},\mathbf{L}^{\mathbb{Z}}_{t-k+1:t},\mathbf{L}^{\mathbb{Z}}_{t-k+1:t}), \nonumber\\
    \mathbf{S}^{'\mathbb{D}}_{t-k+1:t} &= \text{MultiHead}(\mathbf{S}^{\mathbb{D}}_{t-k+1:t},\mathbf{S}^{\mathbb{Z}}_{t-k+1:t},\mathbf{S}^{\mathbb{Z}}_{t-k+1:t}).\nonumber
\end{align}

Similarly to TAM, we can update all modalities by  $\mathbb{D}_t$ and $\mathbb{Z}_t$ to obtain a new representation $\mathbb{D}_t^{\text{CAM}}$ of the sliding window.
\begin{align}\nonumber
     \mathbb{D}_t^{\text{CAM}} & = \text{CAM} (\text{Norm}(DI_{1} + \text{Mask}(\text{TAM}(DI_{1}))), \mathbb{Z}_t) \\\nonumber
     &=(\mathbf{M}^{'\mathbb{D}}_{t-k+1:t},\mathbf{L}^{'\mathbb{D}}_{t-k+1:t},\mathbf{S}^{'\mathbb{D}}_{t-k+1:t}).
\end{align}
where $DI_{1}=\text{Norm}(\mathbb{D}_t+\mathbb{D}_t^{\text{SAM}})$. This updated representation is then passed through the feed-forward and normalization layers to generate the final reconstruction representation.

\subsubsection{Loss Function}
For an anomaly algorithm deployed to production, given a sliding window $\mathbb{D}_t = \{G_{t-k+1},\cdots,G_{t}\}$ with $k$ MST graphs, our primary concern is whether the service instances in the last graph $G_t$ is abnormal or not. Therefore, we only calculate the distance between the input $G_t$ and the output $G_t^{'}$ as
\begin{align}
    \| G_t - G_t^{'} \| = \mathbf{R}\circ \mathbf{R}\nonumber
\end{align}
where $\circ$ indicates the element-wise product, $\mathbf{R}=\text{Concat}$ $((\mathbf{M}_{t} - \mathbf{M}_{t}^{'}), (\mathbf{L}_{t} - \mathbf{L}_{t}^{'}), \mathbf{W}(\mathbf{S}_{t} - \mathbf{S}_{t}^{'}))\in\mathbb{R}^{N\times (F_m+F_l+F_s)}$, and $\mathbf{W}$ is a learnable parameter that maps the third item to the same dimension as the first two items.
Using this distance, we can make predictions for the service instances within an MST graph via supervised learning (i.e., the probability of being abnormabl $P_t\in\mathbb{R}^{N\times 2}$) and unsupervised learning (i.e., the reconstruction error $RE_t\in\mathbb{R}^{N}$), as follows:
\begin{align}
    RE_t&=\text{sum}(\| G_t - G_t^{'} \|,2), \nonumber\\
    P_t &= \text{Softmax}(\text{MLP}(\| G_t - G_t^{'} \|)), \nonumber
\end{align}
where $\text{sum}(\cdot,2)$ indicates the summation by colmuns.

MSTGAD aims to train an efficient anomaly detection model based on limited labeled data and a substantial amount of unlabelled data.
To this end, we adopt a semi-supervised reconstruction loss and a supervised classification loss, which can be combined as follows:
\begin{equation}\label{eq:loss}
    \mathcal{L}oss = \frac{1}{epoch}\mathcal{L}_1 + (1- \frac{1}{epoch})\mathcal{L}_2,
\end{equation}
where $\mathcal{L}_1$ is the loss for data reconstruction, $\mathcal{L}_2$ is the loss for classification, and $epoch$ represents the current epoch number. More concretely, the semi-supervised loss $\mathcal{L}_1$ is defined as:
\begin{equation}
    \mathcal{L}_1 = \frac{1}{m+n}(\eta\sum_{i=1}^m RE_i + \sum_{i=1}^{n_a} \frac{1}{RE_i} + \sum_{i=1}^{n_n} RE_i),
\end{equation}
where $m$ and $n$ are respectively the numbers of unlabeled and labeled service instances in the training data, $RE_i$ indicates the reconstruction error for the $i$-th service instance, $\eta$ is a hyperparameter that balances the weight between labeled and unlabeled data, $n_a$ and $n_n$ are the numbers of abnormal and normal service instances in the training data, respectively, and $n=n_a+n_n$. This semi-supervised reconstruction loss helps the model learn the underlying patterns in the data and detect anomalies based on deviations from these patterns.

The supervised classification loss, on the other hand, helps the model to classify the service instances as normal or abnormal based on the learned patterns. Specifically, we employ the binary cross entropy loss:
\begin{equation}
    \mathcal{L}_2 = -\frac{1}{n}\sum_{i=1}^n [\frac{n_n}{n_a} y_i\log(\hat{y}_i)+(1-y_i)\log(1-\hat{y}_i)],
\end{equation}
where $\hat{y}_i$ denotes the predicted probability for the $i$-th service instance, $\frac{n_n}{n_a}$ is introduced to control the trade-off between normal and abnormal data in the loss function. This loss function enables the model to learn to distinguish between normal and abnormal service instances more accurately, which leads to better performance in detecting anomalies.

\begin{algorithm}[t]
    \caption{The MSTGAD training algorithm}
    \label{alg:MSTGAD}
    \KwIn{Dataset used for training $\{\mathbb{D}_t\}_{t=1}^T$, Encoder ${E}$, Decoder $D$,  Iteration limit $M$;}
    \KwOut{Model Coefficients}

    \textbf{Initialization}: Initialize the weights of $E$ and $D$\;
    $i \leftarrow 1$\;
    \Repeat{$i>M$}
    {
        \For{$t=1$ to $T$}
        {
            $EI_0^t, DI_0^t \gets \text{InputEmbedding}(\mathbb{D}_t)$\;
            $\mathbb{Z}_t \gets E(EI_0^t)$ \;
            $G_t^{'} \gets D(DI_0^t, \mathbb{Z}_t)$\;
            Calculate $RE_t$ and $P_t$\;
            $\mathcal{L} = \frac{1}{i}\mathcal{L}_1 + (1- \frac{1}{i})\mathcal{L}_2$\;
            Update the weights of $E$ and $D$ using $\mathcal{L}$\;
        }
        $i \leftarrow i+1$\;
    }
\end{algorithm}

Note that in Eq.~(\ref{eq:loss}) the weight of the two losses changes with the number of epochs to balance the relative importance of the reconstruction loss and the classification loss during the training process. In the early stages of training, the model relies more on the reconstruction loss to learn the underlying patterns in the data and reconstruct the input data accurately. At this stage, the importance of the classification loss is relatively smaller, as the model has not yet learned to distinguish between normal and abnormal service instances accurately. As the training progresses, the model becomes better at reconstructing the input data and learning the underlying patterns. At this stage, the importance of the classification loss increases, as the model needs to learn to distinguish between normal and abnormal service instances more accurately. Therefore, we adjust the weight of the two losses with the number of epochs, starting with a higher weight for the reconstruction loss and gradually decreasing it while increasing the weight of the classification loss. This approach ensures that the model can learn the underlying patterns in the data effectively while also accurately identifying anomalies, ultimately leading to a more robust and accurate anomaly detection model. The overall training processes of MSTGAD are summarized in Algorithm \ref{alg:MSTGAD}.

\section{Evaluation}
\subsection{Datasets}
To evaluate the effectiveness of our proposed method, we used two multi-modal anomaly detection datasets in our experiment. Both datasets contain metrics, logs, and traces.

\subsubsection{MSDS}
The multi-modal dataset MSDS\footnote{https://zenodo.org/record/3549604}\cite{Dataset:MSDS} is composed of distributed traces, application logs, and metrics from a complex distributed system (Openstack) that is used for AI-powered analytics. The metrics data contains information from 5 physical nodes in the infrastructure, each containing 7 metrics such as RAM and CPU usage. The log files are distributed across the infrastructure and recorded for each node, with a total of 23 features. The trace information encompasses various attributes such as host, event name, service name, span ID, parent ID, and trace ID.
Notably, the MSDS also provides a JSON file containing the ground-truth information for the injected anomalies, including their start and end times, as well as their corresponding anomaly types. Hence, by leveraging this JSON file, the service instances' label information can be easily extracted and assigned within an MST graph. The ratio of normal and abnormal data is approximately 80:1.

\subsubsection{AIOps-Challenge}
The AIOps-Challenge \footnote{https://aiops-challenge.com/} dataset serves as the foundation for the CCF International AIOps Challenge organized by CCF (China Computer Federation), Tsinghua University, and CCB (China Construction Bank) in 2022.
This dataset is derived from a simulated e-commerce system operating on a microservice architecture, with 40 service instances deployed across 6 physical nodes. Each service instance records metrics, encompassing a total of 56 metrics, of which 25 are utilized in this study, including metrics related to RAM and CPU usage.
Additionally, log files are recorded for each service instance, containing a collective set of 5 features, such as timestamps and original logs, among others. The traces within the dataset capture scheduling information among service instances, including timestamps, types, status codes, service instance names, span IDs, parent IDs, and trace IDs.
Within the AIOps-Challenge dataset, three levels of anomalies are intentionally injected, specifically at the service, pod (service instance), and node levels. The injected anomalies are accompanied by their start times, levels, service names, and types. Accordingly, the label information for service instances can be extracted based on the injected anomalies. In specific terms, if a service anomaly is injected, the corresponding service instances are labeled as abnormal. Similarly, if a node anomaly is injected, the service instances configured within that node are all labeled as abnormal. It is crucial to note that the end time of an injected anomaly is not provided. To address this limitation, we manually set the end time with a maximum interval of two minutes. The ratio of normal data to abnormal data in this dataset is approximately 120:1.

\subsection{Experiment Settings}
\label{section:experimentsettings}
Unless otherwise specified, for all datasets, we allocated 60\% of the data for training, 10\% for validation, and the remaining 30\% for testing. Since PLELog, HADES, and our approach MSTGAD are semi-supervised, we randomly selected 50\% of the training data as unlabelled. To evaluate the performance of all approaches, we utilized precision (PR), recall (RC), F1-score, average precision (AP), and area under the receiver operating characteristic curve (ROC/AUC) as the criteria \cite{DeepTraLog:ICSE22,AnomalyDetection:survey:2023, AIOps:survey:21}. 
PR indicates how many of the anomalous events predicted are actual.
RC denotes the percentage of predicted abnormal events versus all abnormal events.
The F1-score is the harmonic mean of precision and recall.
AP indicates the weighted mean of precisions achieved at each threshold, with the increase in recall from the previous threshold used as the weight.

All experiments are run on a Linux server equipped with an Intel(R) Xeon(R) Gold 5318Y CPU, 64 GB RAM, RTX 3090 with 24GB GPU memory, and Ubuntu 18.04.6 OS. Our implementation of MSTGAD is created using Python 3.8.13, PyTorch 1.12.0~\cite{paszkepytorch2019} with CUDA 11.3, and PyTorch Geometric Library 2.2.0~\cite{feyfast2019}. The encoder and decoder layers were set to two, the batch size was 50, the window size was 10, and the Dropout was set to 0.2. We used the AdaBelief~\cite{zhuangadabelief2020} optimizer with an initial learning rate of 0.001 and a step-scheduler with a step size of 0.9 for training. We trained models for up to 300 epochs and utilized early stopping with a patience of 15.

\subsection{Benchmark Algorithms}
To verify the effectiveness of our proposed method MSTGAD, we conducted a comparative analysis with state-of-the-art models for multivariate time-series anomaly detection, including 
TraceAnomaly \cite{TraceAnomaly:ISSRE20}, PLELog \cite{PLELog:ICSE2021}, TranAD \cite{TranAD:VLDB22}, USAD \cite{USAD:KDD20}, SCWarn \cite{SCwarn:ESECFSE21}
and HADES \cite{HADES:ICSE23}. Table \ref{tab:cmpmodels} presents an overview of the distinguishing features of these models. We utilized the open-source codes provided by their respective publications for comparisons.
For these approaches, the window size is 10, the batch size is 50, and the other hyperparameters of them are tuned by grid searching based on the validation set.
\begin{table}[t]
    \centering
        \caption{Characteristics of the comparing models}
        \label{tab:cmpmodels}
        \begin{tabular}{p{2cm} p{2.45cm} cc} \hline
            Method  & Data Used & Supervised Type & Online\\ \hline 
            TraceAnomaly & Trace & Unsupervised & $\times$ \\
            TranAD  & Metric & Unsupervised& \checked\\
            USAD  & Metric & Unsupervised& \checked\\
            SCWarn  & Metric \& Log  &Unsupervised & \checked\\
            PLELog  & Log & Semi-supervised & \checked\\
            HADES  & Metric \& Log & Semi-supervised & \checked\\
            MSTGAD &  Metric\&Log\& Trace & Semi-supervised & \checked \\ \hline
        \end{tabular}
\end{table}

\subsection{Experimental Result}
\subsubsection{Results of Anomaly Detection}
\begin{table}[t]
    \centering
    \caption{Experimental Results on MSDS Dataset}
    \label{tab:results:MSDS}
    \begin{tabular}{p{2cm} p{0.85cm}<{\centering} p{0.85cm}<{\centering}p{0.85cm}<{\centering}p{0.85cm}<{\centering}p{0.85cm}<{\centering}} \hline
    Method  &  PR	& RC	& AUC &	AP &	F1\\ \hline 
    TraceAnomaly &  0.903	&\textbf{0.989}	&0.986	&0.894	&0.944 \\ 
    PLELog & 0.753	&0.663	&0.826	&0.516	&0.705 \\
    TranAD & 0.772	&0.815	&0.907	&0.815	&0.793 \\
    USAD & 0.481	& 0.463	&0.730	& 0.611	&0.472 \\
    SCWarn & 0.440	& 0.371	&0.679	& 0.581	&0.402 \\
    HADES &  0.908	&0.895	&0.947	&0.814	&0.901 \\
    MSTGAD & \textbf{0.946} &0.969 &\textbf{0.996} &\textbf{0.971} &\textbf{0.957} \\\hline
    
\end{tabular}
\end{table}

The average experimental results for the benchmark algorithms on the MSDS and AIOps-Challenge datasets are presented in Table~\ref{tab:results:MSDS} and~\ref{tab:results:Ali} respectively.
Based on the outcomes of the experiments, we can make the following observations:
 
MSTGAD, our proposed method, achieves the best performance among all compared approaches and outperforms all baselines by a significant margin, attaining $F_1$-score of 0.957 and 0.965 on the two datasets. The high scores of MSTGAD indicate that there are few instances of missed anomalies or false alarms, which highlights the effectiveness of our approach for anomaly detection.
\begin{table}[t]
    \centering
    \caption{Experimental Results on AIOps-Challenge Dataset}
    \label{tab:results:Ali}
    \begin{tabular}{p{2cm} p{0.85cm}<{\centering} p{0.85cm}<{\centering}p{0.85cm}<{\centering}p{0.85cm}<{\centering}p{0.85cm}<{\centering}} \hline
    Method  &  PR	& RC	& AUC &	AP &	F1\\ \hline 
    TraceAnomaly & 0.857	& 0.239	& 0.619	& 0.234	& 0.374\\
    PLELog & 0.750	& 0.173	& 0.586	& 0.142	& 0.281 \\
    TranAD & 0.661	& 0.789	& 0.827	& 0.738	& 0.719\\
    USAD   & 0.667	& 0.750	& 0.813	& 0.728	& 0.706 \\
    SCWarn & 0.878	& 0.625	& 0.798	& 0.762	& 0.730 \\
    HADES  & 0.911  & \textbf{0.937}	& 0.958 & 0.865 & 0.924 \\
    MSTGAD & \textbf{1}	    & 0.933	&\textbf{0.974}	& \textbf{0.977} & \textbf{0.965} \\ \hline
    \end{tabular}
\end{table}

Moreover, compared with single-modal approaches,
MSTGAD achieves superior performance. Previous studies have shown that metrics, logs, and traces can all reflect anomalies, and none of them are sufficient when acting alone. Thus, constructing the model based on a single data modality can omit important information hidden in the other data modalities, resulting in performance degradation. Additionally, the log-based approach PLELog's performance may be influenced by the distribution of logs, i.e., the significant difference in the number of logs in different sliding windows.

On the other hand, compared with multi-modal based approaches SCWarn and HADES, MSTGAD outperforms them across all measurements on average. The possible reasons are summarized from three aspects: 1) SCWarn and HADES only use two types of data modalities, i.e., metrics and logs. 2) SCWarn is an unsupervised method and does not use supervised information to guide the training process. Furthermore, it detects anomalies on each modality individually and fails to identify the hidden correlations among the multi-modal data. 3) HADES is designed with an attention-based module, which fuses information between metrics and logs. However, it only detects anomalies for a single service instance and ignores the interaction between multiple service instances.

In contrast, MSTGAD detects anomalies via multi-modal learning based on metrics, logs, and traces simultaneously. The three modalities are effectively integrated via an MST graph, while spatial and temporal attention modules are proposed to model the interactions between different data modalities, and the complementary information among them and specific information are thoroughly fused and effectively integrated.

In conclusion, MSTGAD effectively detects abnormal patterns in all datasets, significantly proving its superiority over all baselines in terms of every evaluation criterion.

\begin{table}[t]
    \centering
    \caption{Experimental Results of Ablation Study on MSDS Dataset}
    \label{tab:results:AblationStudy}
    \begin{tabular}{p{2.1cm}p{0.8cm}<{\centering}p{0.8cm}<{\centering}p{0.85cm}<{\centering}p{0.85cm}<{\centering}p{0.85cm}<{\centering}} \hline
    Method  &  PR	& RC	& AUC &	AP &	F1\\ \hline 
    MSTGAD-Metric &  0.920 	&0.957 	&\textbf{0.996} 	&0.960 	&0.937  \\ 
    MSTGAD-Log & 0.758 	&0.917 	&0.986 	&0.949 	&0.830 \\
    MSTGAD-Trace & 0.720 	&0.920 	&0.977 	&0.876 	&0.807 \\
    MSTGAD-TAM &  0.932 	&0.917 	&0.960	&0.948 	&0.921  \\
    MSTGAD-SAM &  \textbf{0.965} 	&0.929	&0.975	&0.967	&0.947 \\
    MSTGAD-C &0.939	&	0.929		& 0.967	&	0.955 &	0.933 \\
    MSTGAD & 0.946 &\textbf{0.969} &\textbf{0.996} &\textbf{0.971} &\textbf{0.957}\\\hline
    \end{tabular}
\end{table}
\subsubsection{Ablation Study}
\label{sec:ablationstudy}
To comprehensively assess the efficacy of different modules of MSTGAD, we conducted an ablation study on the MSDS dataset, and the experimental results are listed in Table~\ref{tab:results:AblationStudy}, where \textdbquote{MSTGAD-X} indicates that MSTGAD is executed without the module or data modal \textdbquote{X}. For example, \textdbquote{MSTGAD-Metric} means MSTGAD is executed without the metrics data.
It can be seen from Table~\ref{tab:results:AblationStudy} that all modules and data modalities contribute significantly to improving the performance of MSTGAD.
Concretely, MSTGAD exhibits a significant improvement over MSTGAD-Metric, MSTGAD-Log, and MSTGAD-Trace with an impressive margin of 3.933\% in terms of $F_1$-score on average, showcasing its exceptional ability to effectively utilize multi-modal data for anomaly detection. Notably, MSTGAD-Trace performs worse than MSTGAD-Metric and MSTGAD-Log. Recall that in MSTGAD, both metrics and logs serve as node features, while only traces are utilized as edge features. When trace data is removed, the SAM module can only utilize the node features, compromising the effectiveness of updating and enhancing the graph representation. Consequently, the complementary information from different modalities cannot be effectively integrated, resulting in suboptimal performance. Remarkably, similar observations have been reported in Eadro~\cite{Eadro:ICSE23:multimodal}, further emphasizing the fundamental role of trace data in combining and exploiting information from the other two modalities.


In the meantime, MSTGAD demonstrates statistically significant or, at the very least, comparable performance compared to MSTGAD-TAM and MSTGAD-SAM. The impressive performance of MSTGAD validates the effectiveness of incorporating both spatial and temporal attention mechanisms in anomaly detection. Moreover, MSTGAD-SAM outperforms MSTGAD-TAM, indicating that modeling temporal dependencies is crucial and highly effective in addressing anomaly detection challenges.

MSTGAD-C indicates that MSTGAD runs without modeling the common temporal dependency shared by different modalities, i.e., the matrix $\mathbf{C}$ is removed from Eq.(\ref{eq:temporalAtt}).
As shown in Table~\ref{tab:results:AblationStudy}, MSTGAD achieves a better performance than MSTGAD-C.
It demonstrates the effectiveness of capturing the common temporal dependencies shared by the multi-source data and fusing the complementary information among them.

 
\subsubsection{Parameter Sensitivity Analysis}
\begin{figure}[t]
\centering
        \subfigure[Number of layers of encoder and decoder]{
        \includegraphics[width=0.22\textwidth,height=1.2in]{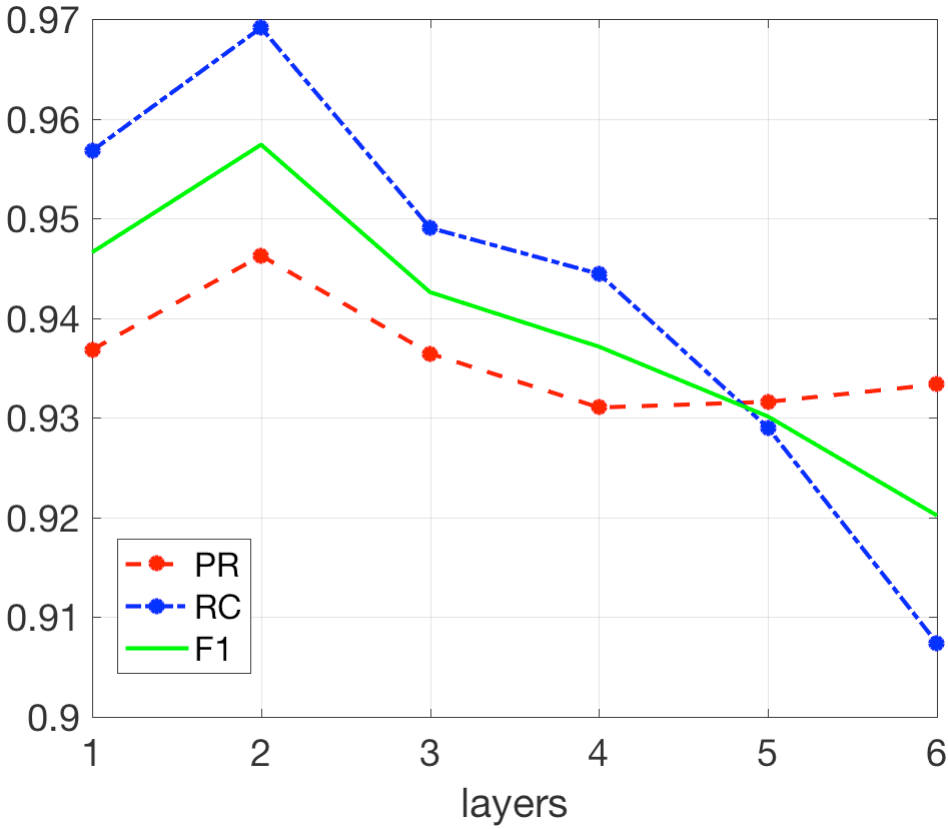}}\hspace{.2in}
        \subfigure[Balance weight $\eta$]{\includegraphics[width=0.22\textwidth,height=1.2in]{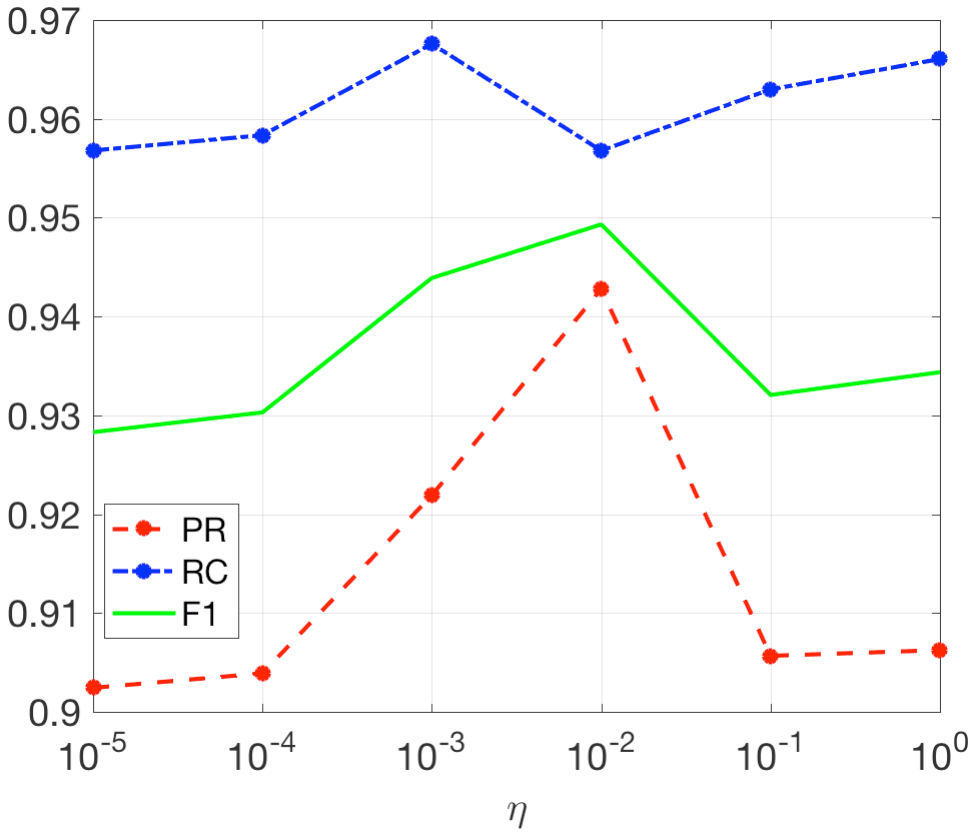}} 
        
        \subfigure[Percent of labelled data]{
        \includegraphics[width=0.23\textwidth,height=1.25in]{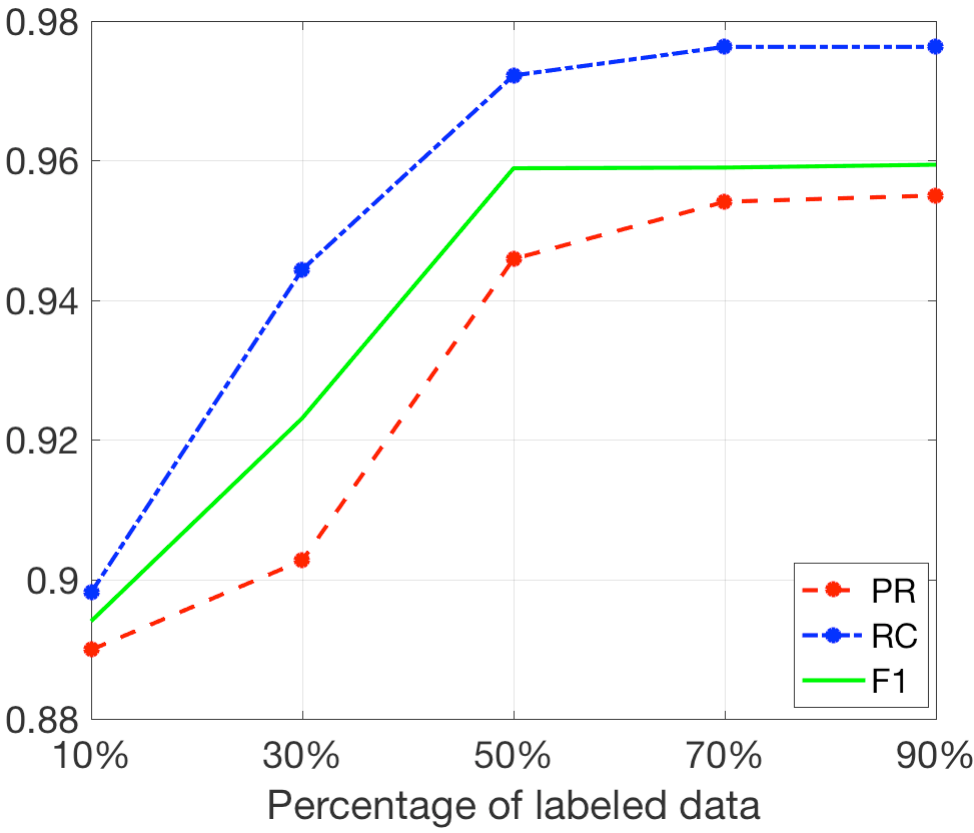}} 
        \subfigure[Sliding window size]{
        \includegraphics[width=0.23\textwidth,height=1.25in]{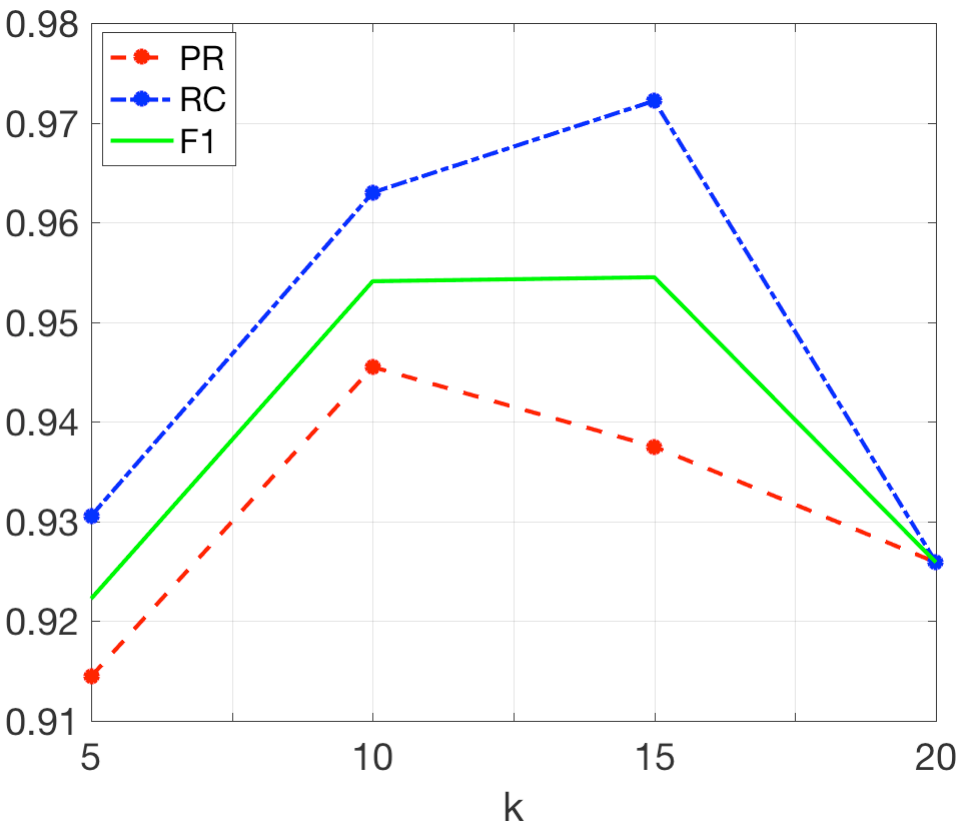}} 
        
        \caption{Parameter Sensitivity Analysis}
        \label{fig:para}
\end{figure}

The proposed method, MSTGAD, incorporates several crucial hyperparameters, including the sliding window size $k$, the number of encoder and decoder layers, the balance weight $\eta$, and the percentage of labeled data. To evaluate the sensitivity of these parameters, we conducted experiments on the MSDS dataset, and the results are displayed in Fig. \ref{fig:para}.

Fig. \ref{fig:para}(a) demonstrates the impact of the number of encoding and decoding layers on the performance of MSTGAD. We observed that a smaller or larger number of layers typically leads to lower performance, with the optimal range being between two and six layers. The performance of MSTGAD decreased with an increase in the number of hidden layers, probably due to the over-smoothing that makes the features indistinguishable, leading to reduced classification accuracy.

Fig. \ref{fig:para}(b) illustrates the impact of the balance weight $\eta$ on the precision, recall, and $F_1$-score of MSTGAD. As described in Section \ref{sec:MSTGAD}, $\eta$ controls the trade-off between labeled and unlabeled data in the reconstruction loss. The results indicate that both precision and $F_1$-score decrease with an increase in $\eta$, primarily due to the model giving more attention to the noisy unlabeled data points. $\eta$ is typically set between $10^{-4}$ and $10^{-1}$, based on the preference for precision and recall.

The impact of the percentage of labeled data for the training stage on the performance of MSTGAD is depicted in Fig.~\ref{fig:para}(c). The results demonstrated that the performance of MSTGAD significantly improved with an increase in labeled data. It is worth noting that most of the labeled data are normal patterns that are easily obtainable in real-world applications.

Finally, Fig.~\ref{fig:para}(d) shows the impact of the sliding window size $k$ on the precision, recall, and $F_1$-score of MSTGAD. Smaller or larger values of $k$ typically lead to lower performance due to insufficient or unrelated features in the window. We observed that the best performance was achieved when $k$ was set to 10, as shown in Fig.~\ref{fig:para}(d).

\begin{table}[t]
        \centering
        \caption{Statistical characteristics of the experimental datasets and the processing time of MSTGAD}
        \label{table:timestatistic}
        \begin{tabular}{ccccc } \hline
        \multirow{2}{*} {Dataset} & \multirow{2}{*}{ Interval}	& \# Metrics & \multirow{2}{*}{\# Logs} & \multirow{2}{*}{\# Spans} \\
         &  	& (instance*metric) &  &  \\\hline 
        MSDS &  1s	& 5*3 & 11 & 152 \\ \hline
        AIOps &  1min  &40*25 & 9260 & 6253 \\
        \hline

        \multirow{2}{*} {Dataset} & \multicolumn{4}{c}{Running Time} \\\cline{2-5}
        & Preprocessing & Constructing graph & Detecting & Total \\ \hline
        MSDS & 0.014s & 0.0088s & 0.0035s & 0.0263s  \\
        AIOps&  6.7946s & 0.173s & 0.4648s & 7.4324s \\ \hline
    \end{tabular}
\end{table} 


\subsubsection{Real-time Detection}
In the proposed method, the three modalities are first extracted and then integrated using an MST graph structure for anomaly detection.
It is essential for the time consumption of this step to be smaller than the process interval of a microservice system, ensuring real-time detection capabilities. In this section, we present experimental results conducted on the MSDS and AIOps-Challenge datasets to demonstrate the efficiency of our method. The summarized results are presented in Table \ref{table:timestatistic}, where \# Metrics, \# Logs, and \# Spans correspond to the average number of metrics, logs, and spans observed within each processing interval. It is important to note that all experiments were conducted exclusively using CPU resources, without utilizing GPU or multithreading techniques for acceleration. The results unequivocally illustrate the applicability of the proposed method to real-world applications, exhibiting its capability to perform real-time anomaly detection.

\section{Conclusion}
In this paper, we propose a novel semi-supervised graph-based anomaly detection method MSTGAD that utilizes attentive multi-modal learning. 
We fuse the multi-modal data and represent them via an MST graph, where each node corresponds to a service instance, and the edge indicates the scheduling relationship between different service instances. Based on the MST graph sequences, we construct a transformer-based neural network with both spatial and temporal attention mechanisms to automatically and accurately detect anomalies in a timely manner. Experimental results demonstrate that MSTGAD achieves superior performance compared to state-of-the-art approaches. Furthermore, our results verify that effectively modeling the correlation between different data modalities and the time dependency within each data modality can further improve the performance of anomaly detection.

\bibliographystyle{IEEEtran}
\bibliography{reference}
\vspace{12pt}

\end{document}